# Looking at discourse in a corpus: The role of lexical cohesion[1]

Berber Sardinha, Tony
Catholic University of Sao Paulo, Brazil
tony4@uol.com.br
www.tonyberber.f2s.com

**Abstract**

This paper is aimed at reporting on the development and application of a computer model for discourse analysis through segmentation. Segmentation refers to the principled division of texts into contiguous constituents. Other studies have looked at the application of a number of models to the analysis of discourse by computer. The segmentation procedure developed for the present investigation is called LSM ('Link Set Median'). It was applied to three corpus of 300 texts from three different genres. The results obtained by application of the LSM procedure on the corpus were then compared to segmentation carried out at random. Statistical analyses suggested that LSM significantly outperformed random segmentation, thus indicating that the segmentation was meaningful.

**1 Introduction**

Although there are several papers looking at the use of computers in linguistic research (e.g. McEnery & Wilson, 1996), there is a lack of computer-aided or corpus-based studies looking at how texts are organised (Scott, 1997). One reason for this is that there is also a shortage of computerisable models of text organisation. One such model is offered by Hoey (1991), which describes how texts hang together by means of lexical cohesion.

The aim of this paper is to apply Hoey's (1991) model of patterns of lexical cohesion in text to text segmentation texts. Segmentation here means using the computer to divide a written text into acceptable parts (or segments). An acceptable segment is one which conforms to a certain standard, and in the case of this investigation, it means matching the section divisions placed by the authors of the text. Several segmentation procedures have been reported in the literature (Hearst, 1993, Morris, 1988; Okumura & Honda, 1994; see Berber Sardinha 1997 for more references). While these systems are computationally robust, most of them are not theoretically sound from the point of view of discourse analytic research. Crucially, they neglect to take into account how sentences ('miniature packages of information', according to Hoey, 1991) or clauses create texture by relating to each other lexically.

I intend to show that integration and segmentation are two manifestations of the same phenomenon: 'all texts are about difference and sameness' (Hoey, personal communication); difference surfaces as segments, while sameness is made evident by the existence of meaningful multiple repetition (bonding).

---

[1] I am grateful to Mike Scott, Geoff Thompson, and Chris Butler for reading and commenting on earlier versions of this paper.

The kinds of segments at which this study aims are lexical segments, or contiguous portions of written text consisting of at least two sentences held together by lexical cohesive links.

## 2 Links

The system of analysis proposed by Hoey to describe lexical patterns in text is based on the notion of link. A link occurs whenever there is a repetition of an item in two separate sentences. The term 'link' does not imply directionality (Hoey, 1991, p.52) thus allowing for the creation of webs among lexical items, unlike the traditional notion of 'tie'.

## 3 Patterns of lexis and text segmentation

In view of the lack of existing segmentation procedures which take into account the way lexis operates in text, a procedure known as 'Link Set Median' (LSM) was devised.

Hoey's model is based on the analysis of links between pairs of sentences. To place segments in text, it would be better to look at the similarity between all the sentences with which each adjacent sentence shares lexical items. This might provide some indication of similarity between two sentences even in cases where there are no lexical links shared between the adjacent pair. To achieve this, the concept of link set must be introduced. The set of sentences with which each sentence has links can be seen to form a link set. For instance, if sentence 1 has three links with sentence 6 and two links with sentence 4, then the link set for sentence 1 would be {4,4,6,6,6}, that is, the number 4 is entered twice, one for each link with sentence 4, and the number 6 is entered three times, one for each link with sentence 6. Given that cohesion is a measure of topic shifts (Hoey, 1991) and segmentation (e.g. Hearst, 1993), the simplest measure of where the cohesion is would be to see every cohesive item as a measure of similarity between two sentences. Lexical cohesion is a measure of similarity (Hoey, 1991), and therefore similarity can be assessed by looking at the lexis shared among sentences. Since each lexical item is a separate measure of similarity, if there are three lexical items shared there are three points of similarity, hence the similarity can be recorded three times. The notion of link set as a record of similarity is therefore convenient in that it enables the researcher to observe the degree of similarity between two sentences.

## 4 Summary of the LSM procedure

The notion of link sets was built into a system for segmenting written texts in English known as LSM - Link Set Median. The main procedures involved in segmenting texts with LSM are listed below:

1. Identify the links for all sentences in the text
2. Create the link sets
3. Compute the median for each link set
4. Calculate the median difference for each link set and its predecessor
5. Compute the average (mean) median difference for the text

6. Compare each (link set) median difference to the (text) average median difference
7. If the median difference is higher than the average, insert a segment boundary
8. Locate the section boundaries in the text and disregard sections starting with sentence 1
9. Compare segment and section boundaries

Due to a lack of space in the present article, the reader is referred to Berber Sardinha (1997, pp. 245-261, and 298-308) for further details of the segmentation algorithm.

**5 Data, method, results**

LSM was applied to the task of segmenting three corpora of written texts in English, each containing 100 texts each (totalling 300 texts) of a different genre, namely research articles, business reports, and encyclopaedia articles. The corpus totals 1,262,710 words.

In order to evaluate the performance of the segmentation, two measures commonly found in the Information Retrieval literature will be used: recall and precision. In our case, recall scores refers to the total boundaries correctly inserted by LSM divided by the number of reference segments (existing orthographic section boundaries). Precision, in turn, is the number of section headings divided by the total number inserted by LSM. A perfect score on recall indicates that the procedure has identified all of the reference segments in the text or texts, while a perfect score on precision shows that the procedure has only inserted segment boundaries that matched reference segments.

A central question is whether LSM segmentation is meaningful or arbitrary. In order to rule out the possibility that it is arbitrary we need to compare the LSM segmentation to segmentations of the same texts carried out at random. The results appear in table 1.

| Measure | Method | Research Articles | Business Reports | Encyclopedia Articles |
|---|---|---|---|---|
| Recall | LSM | 22.96% | 21.38% | 17.26% |
|  | Random | 13.64% | 15.92% | 11.11% |
| Precision | LSM | 10.09% | 22.37% | 12.95% |
|  | Random | 4.92% | 13.4% | 8.01% |

Table 1: Comparison of LSM and Random segmentations (All LSM X Random differences are significant at $p < .0001$.)

The figures represent mean percentage values for recall and precision obtained across six link levels. On average, LSM reached 20.5% recall and 15.1% precision, while random segmentation obtained an average of 13.6% recall and 5.4% precision. LSM figures are therefore significantly higher for both recall and precision. LSM achieved higher results in other situations. At link level 2 (at which only those sentences with 2 or more links with any other entered link sets), LSM segmentation achieved over 38% recall (encyclopedia articles) and 27% precision (business reports). In a 25-text sample preliminary study, LSM obtained higher scores: 32% recall and 30% precision (Berber Sardinha, 1997, p.307).

Although low, these figures are well within the range of scores reported in the literature for other procedures (Hearst, 1993; Morris, 1988; Okumura & Honda, 1994). In fact, these score place LSM as the second best performing segmentation procedure in a comparison with three other algorithms (see Berber Sardinha, 1997, p.311). TextTile (Hearst 1993) came first for both recall (37.5%) and precision (100%). The other scores are as follows. For recall, Okumura & Honda and LSM (32%), Morris (31%). And for precision, LSM (29.5%), Okumura & Honda (25%), Morris (24%).

More important than the competition is the possibility of cooperation among procedures, since no segmentation system is perfect. Combining LSM and TextTile (Hearst, 1993) segments, for example, accounted for 80% of the boundaries in a corpus, whereas each one on its own located only about half as many segments (Berber Sardinha, 1997, p.315).

**6 Example**

The text used to illustrate segmentation is an encyclopaedia article on Equatorial Guinea. In the excerpt below there is one segmentation error (sentence 22) and one match (sentence 26). Sentence numbers appear in square brackets, and section headings are shown in italics.

[0017] *Economy and Government*

Agriculture is the main source of livelihood in Equatorial Guinea. [0018] The principal export is cacao, which is grown almost entirely on Bioko. [0019] Coffee is grown on the mainland, which also produces tropical hardwood timber. [0020] Rice, bananas, yams, and millet are the staple foods. [0021] Local manufacturing industries include the processing of oil and soap, cacao, yucca, coffee, and seafood. [0022] The monetary system is based on the franc system (2864 CFA francs equal US $1; 1990). [0023] Under the 1982 constitution, the president was elected by universal suffrage to a seven-year term, and members of the legislature were elected to five-year terms. [0024] The Democratic Party of Equatorial Guinea was the sole legal political party. [0025] A new multiparty constitution was approved in 1991.

[0026] *History*

The island of Fernando Pó was sighted in 1471 by Fernao do Po, a Portuguese navigator. [0027] Portugal ceded the island to Spain in 1778.

    (Source: Encarta 1994)

A remark is in order about the segmentation error at sentence 22. Sentences such as 18, 19, and 20 are loosely connected and could be read in any order. Sentence 22 marks a break in the section since there is a topic shift from 'agriculture and manufacturing' to 'currency'. The lexis in sentence 22 is very different from the preceding sentences (francs, dollars vs manufacturing, foods, etc). Thus, one could argue that sentence 22 is in fact a legitimate segmentation point and so LSM was not entirely wrong.

# 7 Conclusion

The present investigation contributes to current research on lexical cohesion by stressing the role of this type of cohesion as a means of creating topic divisions in texts. There is evidence in this investigation to suggest that writers signal topic shifts (as indicated by the section headings that they place in their texts) through breaks in lexically cohesive clusters. There are several important aspects of the whole procedure which could be improved, and there are discussed at length in Berber Sardinha (1997).